\documentclass[runningheads]{llncs}
\usepackage{graphicx}
\usepackage[caption=false]{subfig}
\usepackage{rotating}
\usepackage{amsfonts}
\usepackage{amsmath}
\usepackage[dvipsnames]{xcolor}
\usepackage{multirow}

\begin{document}

\title{WarpEM: Dynamic Time Warping for Accurate Catheter Registration in EM-guided Procedures}
\titlerunning{WarpEM: DTW for Catheter Registration in EM-guided Procedures}

\author{
Ardit Ramadani\inst{1, 2, 3} \and 
Peter Ewert \inst{2} \and 
\\Heribert Schunkert \inst{2, 3} \and 
Nassir Navab\inst{1, 4, 5} 
} 

\authorrunning{A. Ramadani et al.}

\institute{
Computer Aided Medical Procedures,\\ Technical University of Munich, Munich, Germany\\ 
\email{ardit.ramadani@tum.de}\\
\and German Heart Center Munich, Munich, Germany
\and German Centre for Cardiovascular Research,\\ Munich Heart Alliance, Munich, Germany
\and Munich Institute of Robotics and Machine Intelligence,\\ Technical University of Munich, Munich, Germany
\and Computer Aided Medical Procedures,\\ Johns Hopkins University, Baltimore, USA
}

\maketitle              
%

\begin{abstract}
Accurate catheter tracking is crucial during minimally invasive endovascular procedures (MIEP), and electromagnetic (EM) tracking is a widely used technology that serves this purpose. However, registration between preoperative images and the EM tracking system is often challenging. Existing registration methods typically require manual interactions, which can be time-consuming, increase the risk of errors and change the procedural workflow. Although several registration methods are available for catheter tracking, such as marker-based and path-based approaches, their limitations can impact the accuracy of the resulting tracking solution, consequently, the outcome of the medical procedure. 

This paper introduces a novel automated catheter registration method for EM-guided MIEP. The method utilizes 3D signal temporal analysis, such as Dynamic Time Warping (DTW) algorithms, to improve registration accuracy and reliability compared to existing methods. DTW can accurately warp and match EM-tracked paths to the vessel's centerline, making it particularly suitable for registration. The introduced registration method is evaluated for accuracy in a vascular phantom using a marker-based registration as the ground truth. The results indicate that the DTW method yields accurate and reliable registration outcomes, with a mean error of $2.22$mm. The introduced registration method presents several advantages over state-of-the-art methods, such as high registration accuracy, no initialization required, and increased automation.

\end{abstract}


\section{Introduction}
\label{sec:Intro}
Minimally invasive endovascular procedures (MIEP) are becoming increasingly popular due to their non-invasive nature and quicker recovery time compared to traditional open surgeries. MIEP encompasses various applications, from peripheral artery disease treatments to complex procedures in kidneys, liver, brain, aorta, and heart. Catheter tracking is an essential component of these procedures, allowing for accurate guidance of the catheter through the vasculature~\cite{abi-jaoudeh_multimodality_2012}. 

Electromagnetic (EM) tracking is a widely used technology for catheter tracking in MIEP. However, accurate registration between preoperative images and the EM tracking system remains a challenge~\cite{franz_electromagnetic_2014,ramadani_survey_2022}. Existing registration methods often require manual interaction, which can be time-consuming and may alter the procedural workflow. Achieving accurate and effective registration is essential for a seamless and fast integration of EM tracking, preoperative data, and the patient into the same intraoperative coordinate space.

\subsection{Related work}
\label{sec:RelatedWork}

\subsubsection{Electromagnetic tracking}
is a popular tracking technology that uses integrated sensors in the catheter tip and a field generator to enable localization of the sensor's pose in 3D. EM tracking does not require line-of-sight, which makes it particularly advantageous for MIEP~\cite{zhang_electromagnetic_2006}. The sensors come in various shapes and sizes and can be tracked in translation and orientation, with respect to the generated EM field. Therefore, EM tracking is typically used in conjunction with intra- or preoperative images to guide the procedure~\cite{franz_electromagnetic_2014,ramadani_survey_2022}.  

\subsubsection{Registration} is an essential step that enables intraoperative guidance of procedures through tracking technologies such as EM. Numerous methods have been developed and presented for EM tracking registration, particularly in vascular procedures~\cite{liao_review_2013,matl_vascular_2017}. One method commonly used for registering all involved components in the same intraoperative coordinate space involves the use of external markers or fiducials, which must remain affixed to the patient's body throughout the preoperative and intraoperative phases of the procedure. In order to achieve this registration, surgeons are required to match multiple physical marker points with their corresponding preoperative positions in the image, generating a set of point correspondences that can then be used to calculate the registration transformation. Some of the works using the marker-based registration method are presented in~\cite{lin_automatic_2020,manstad-hulaas_three-dimensional_2011,mittmann_reattachable_2022,wood_navigation_2005}. However, this method is subject to several disadvantages, including changes to the procedural workflow, the necessity of ensuring marker stability between the pre- and intraoperative phase, and sensitivity to natural changes in the patient's anatomy.

Other registration methods include the use of the EM-tracked catheter paths and registration to the vessel's centerline, using iterative closest point (ICP) algorithms, as described in~\cite{luo_bronchoscopic_2014,nypan_vessel-based_2019}, while others explored the registration of EM paths to points within the vasculature, as reported in~\cite{de_lambert_electromagnetic_2012}. Extensive research has been conducted on these methods, which have demonstrated accurate and reliable registration results. However, these methods are also subject to some drawbacks, such as limitations in accurately capturing the movement of the catheter, requiring a good initialization for accurate registration, and reduced accuracy when missing data segments or limited number of data points.

\subsubsection{Dynamic Time Warping} is a widely-used technique in signal processing to compare and align two time-dependent sequences. Dynamic Time Warping (DTW) calculates the similarity between the sequences by optimally aligning them in a nonlinear fashion, taking into account time differences and sequence sampling rates~\cite{muller_dynamic_2007,efrat_curve_2007}. The warping function enables the matching of corresponding features in the sequences, allowing for accurate alignment even when there are temporal differences or missing data points. DTW has many applications, including biomedical signal processing, speech, and gesture recognition~\cite{muller_dynamic_2007,efrat_curve_2007}. 

Despite the well-known advantages of DTW as a technique for signal alignment, its application as a registration method in EM-tracked MIEP is still unexplored. Previous works have reported using DTW and EM tracking; however, these works have primarily used DTW for data processing and evaluation due to its advantages in interpreting intravariability, rather than for registration purposes~\cite{sielhorst_synchronizing_2005,zhang_extracting_2019}. This paper introduces a novel approach to registering EM tracking systems and preoperative images using DTW. To the best of the authors' knowledge, this is also the first work that explores the temporal analysis of 3D EM catheter paths, which are subsequently used to register two systems within a single coordinate space. The introduced method utilizes DTW to warp the time-dependent 3D EM-tracked catheter path to the vasculature's centerline, creating corresponding points between the two. These points are then filtered based on the minimal cost in the DTW algorithm to generate a set of correspondence points that are used for registration between the EM path and centerline.

\section{Method}
\label{sec:Method}

\begin{figure*}[htp]
\centering
    \includegraphics[width=\textwidth]{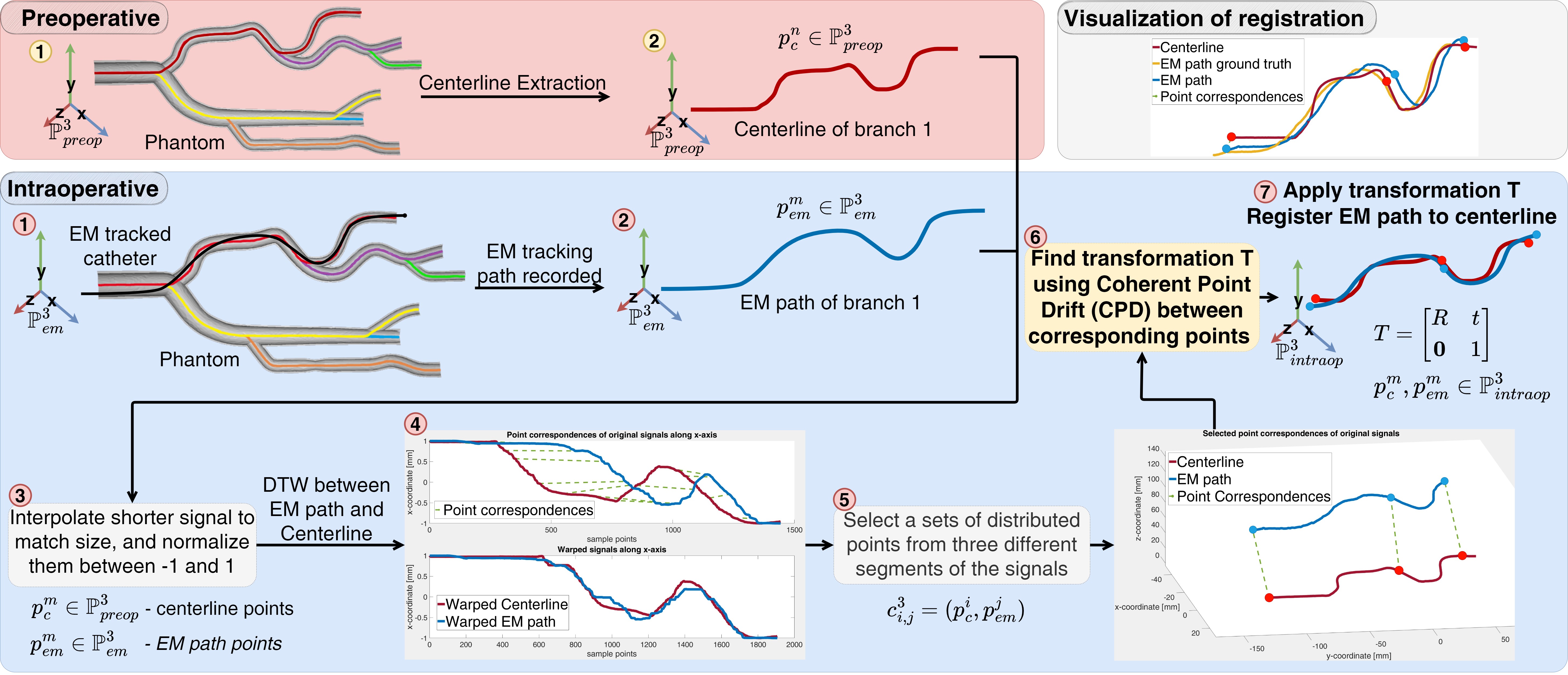}%
    \caption{Overview of the introduced DTW registration method for EM guided MIEP.}
\label{fig:method_overview}
\end{figure*}

This paper introduces a novel approach for automatically registering EM tracking systems to preoperative images using DTW. The introduced method includes a preoperative phase, during which the targeted vasculature is segmented from preoperative images such as magnetic resonance imaging (MRI), computed tomography (CT), or CT angiography (CTA), and the centerlines are extracted using the SlicerVMTK toolkit. In the intraoperative phase, the catheter with an integrated EM sensor in its tip is guided through the respective branch of the vascular tree and records a 3D EM path. In this particular implementation, a catheter-shaped EM sensor is used instead. The recorded EM path is then processed using DTW and warped to the centerline, providing point correspondences between the EM path and the centerline. These correspondences are used to perform a closed-form solution using Coherent Point Drift (CPD) algorithm, translating the EM path to the centerline, resulting in the registration of the two coordinate spaces. A detailed overview of the method is presented in Fig.~\ref{fig:method_overview}.

\subsection{EM tracking system}
\label{sec:EMmethod}

\begin{figure*}[htp]
\centering
    \subfloat[\label{fig:systemsetup}]{\includegraphics[width=0.54\textwidth]{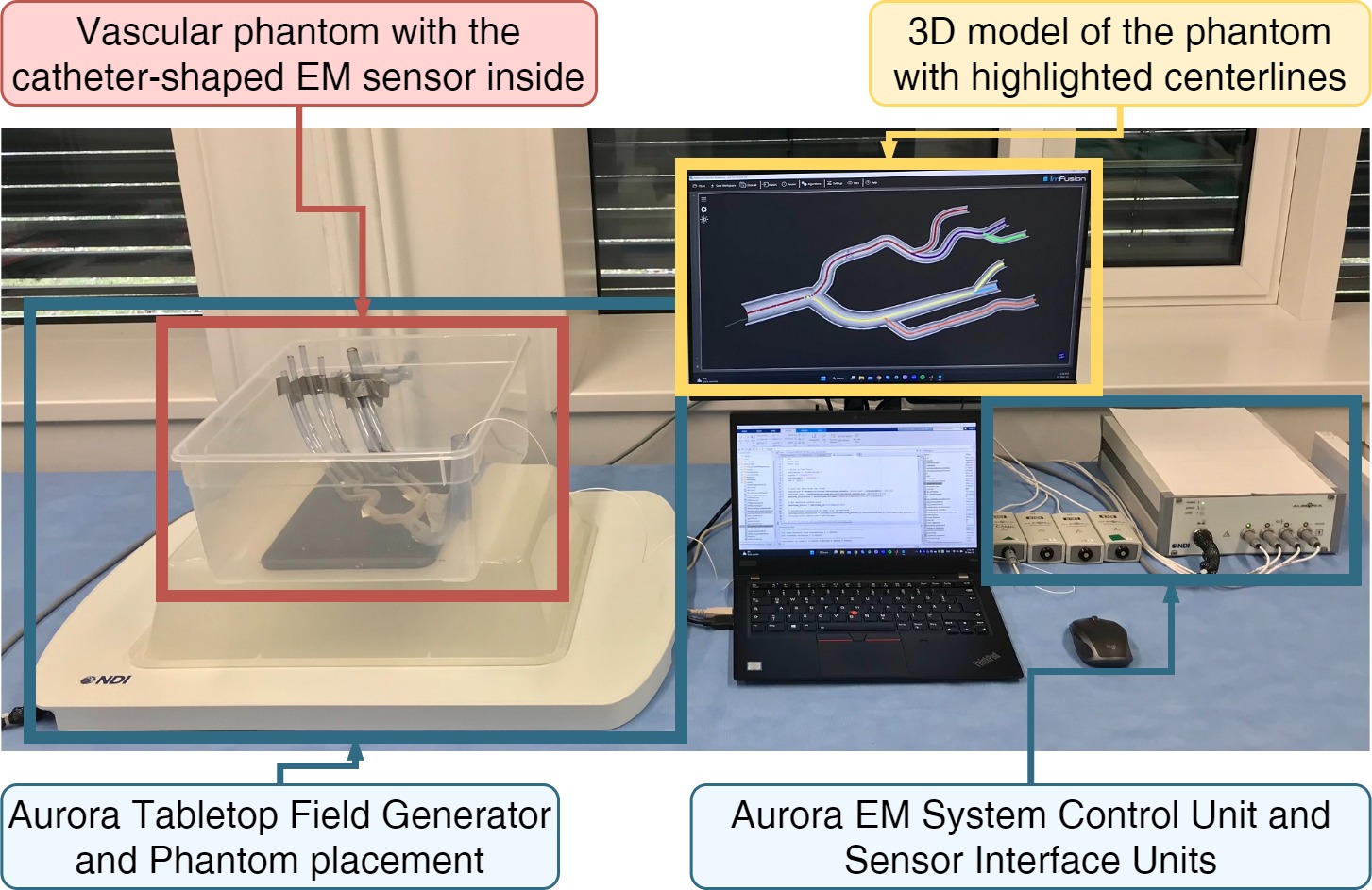}}\hfill
    \subfloat[\label{fig:phantom}]{\includegraphics[width=0.44\textwidth]{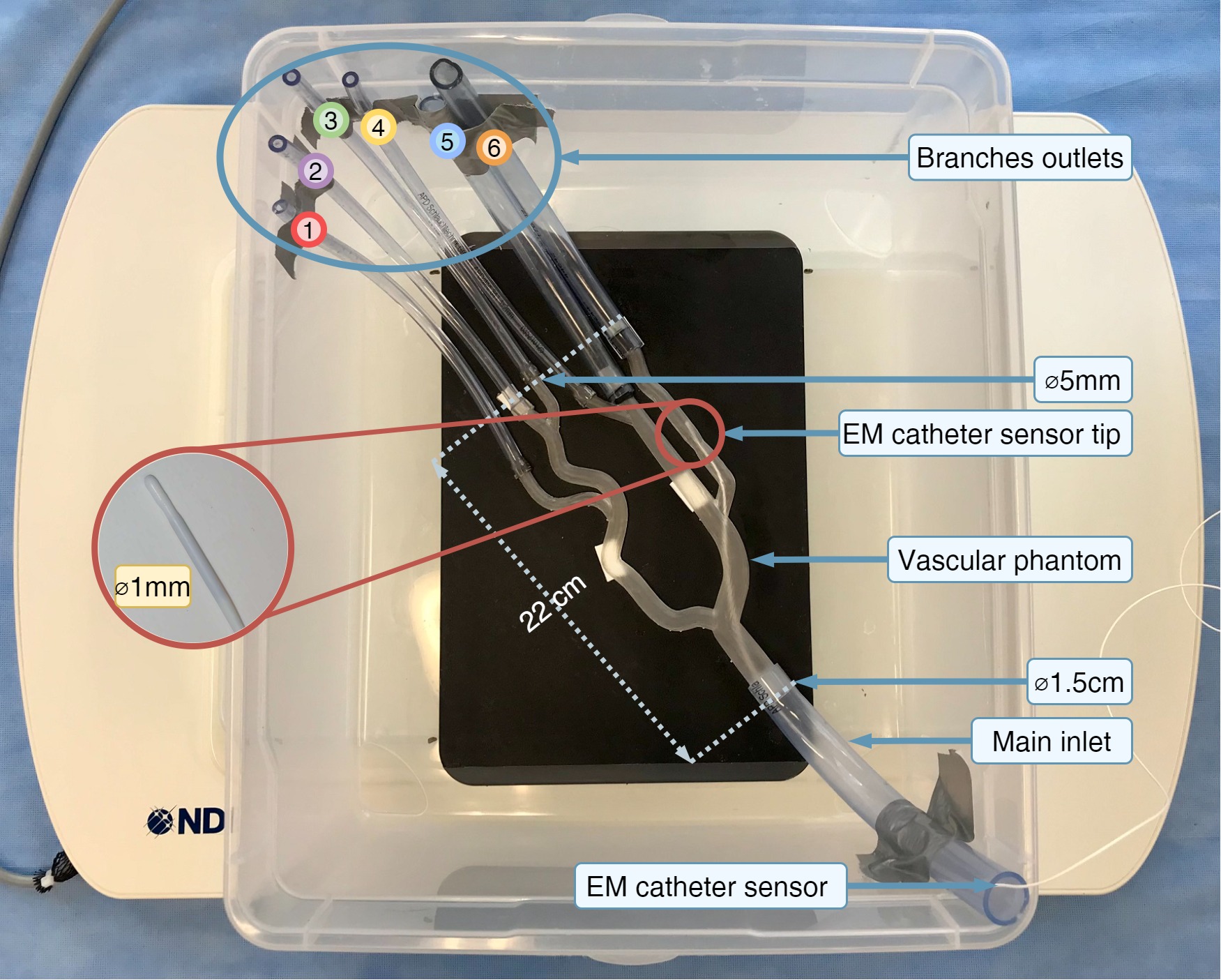}}
    \caption{Full system setup including: (a) EM tracking system with the vascular phantom, EM catheter-shaped sensor, and phantom model visualization on the screen; and (b) Top-view of the vascular phantom used in the experiments, featuring labeled branches for easy identification and color-matching with the results.}
\label{fig:fullsetup}
\end{figure*}

The tracking technology used in this paper is the Aurora tracking system from Northern Digital Inc. - NDI (Waterloo, Ontario, Canada). It comprises of a system control unit, sensor interface unit, and a tabletop field generator, which allows for tracking of EM sensors in a $420$x$600$x$600$ mm space. This tracking system is capable of recording data with a frequency of up to $40$Hz. The Aurora 5DOF FlexTube, which has a $1$mm diameter, was the catheter-shaped EM sensor used in this paper. It is highly versatile in applications since it can be navigated independently or integrated into catheters. The EM sensor was used without a catheter and was navigated in the phantom through its long cable. The tracking data is recorded using ImFusion Suite software (ImFusion GmbH, Munich, Germany), running on a laptop computer with the following specifications: Windows $11$Pro, Intel Core i$7$-$8565$U CPU, $16$GB RAM, and Intel UHD $620$ Graphics. Fig.~\ref{fig:fullsetup} provides a detailed overview of the experimental setup, including the EM tracking system and the phantom utilized in this paper.

\subsection{Phantom}
\label{sec:phantom}

For EM tracking data acquisition purposes of this paper, an STL model obtained from~\cite{Sutton_2020} was utilized. The STL model was simplified while preserving six of the primary branches that represent natural vasculature features, including bifurcations, stenosis, and curvatures. Additionally, the model was resized to dimensions suitable for catheterization, with vessel diameters ranging from $1.5$cm to $5$mm and a length of $22$cm. In order to enable visibility of the catheter-shaped EM sensor from the outside, the phantom was 3D printed using rigid transparent polylactic acid (PLA) material. The phantom was then rigidly fixed in a box and positioned on top of the EM tracking field generator to conduct the experiments and record EM tracking data. Fig.~\ref{fig:phantom} provides a detailed illustration of the phantom and the catheter-shaped EM sensor utilized in this paper.

\subsection{Dynamic Time Warping Registration}
\label{sec:DTWmethod}

During the intraoperative phase, two preoperative components are used, namely the segmented vascular model and its corresponding branch centerline points. The centerline points are referred to as $p_c^n \in \mathbb{P}^3_{preop}$, where $n$ represents the number of points, and $\mathbb{P}^3_{preop}$ represents the 3D preoperative coordinate space. Firstly, the EM catheter-shaped sensor is guided through the vascular phantom, and the resulting EM path is recorded. The EM path points are referred to as $p^m_{em} \in \mathbb{P}^3_{em}$, where $m$ represents the number of points, and $\mathbb{P}^3_{em}$ represents the 3D coordinate space of the EM tracker. In order to use DTW with 3D signals, the number of points must be consistent across the signals; therefore, the signal with fewer points is linearly interpolated to match the number of points in the other signal. Here, the centerlines are interpolated to match the EM paths, $p_c^n \rightarrow p_c^m$.

In the next step of the introduced method, the centerline and the EM path signals are normalized between $-1$ and $1$ to bring the signals into a temporary common coordinate space. The DTW algorithm registration process assumes that the orientation of the phantom (patient) and the preoperative model are similar. Here forwards, DTW decomposes both 3D signals into their respective axes, namely $x$, $y$, and $z$ over time, and matches each point from the EM path to the centerline's counterpart. This iterative process stretches the two signals until the sum of the euclidean distances between corresponding points is minimized. The output warp paths represent the corresponding indices that have been warped from the DTW algorithm, creating a set of minimum cost correspondences between the EM path and centerline, $c^u_{i,j} = (p_{c}^i,p_{em}^j)$. The variable $c$ in the equation refers to the set of corresponding points between the two signals, $u$ represents the total number of correspondences, while $i$ and $j$ represent the indices of the corresponding matched points. For each point in the EM path, there exist one or more points in the centerline that have been warped together and vice versa. For further reading on the DTW algorithm used in this paper, please refer to the following works with more implementation details~\cite{paliwal_modification_1982,sakoe_dynamic_1978}.

In order to register the two signals, we leverage the point correspondences generated by the DTW algorithm to select three sets of equally distributed points from three equal segments of the signals, $c^3_{i,j} = {(p_{c}^i, p_{em}^j)}$. The correspondences in each signal segment are selected based on the minimum cost return function of the DTW algorithm, where the sum of the euclidean distances between corresponding points is minimal. Utilizing the three segments facilitates the equitable distribution of point matching across the entirety of the signal. This step is crucial to ensure high confidence matching and produce a reliable registration that does not rely entirely on one part of the vasculature. The selected correspondence points are then used to find the rigid transformation between them using the CPD algorithm, as described in~\cite{myronenko_point_2010}. The introduced solution is a closed-form algorithm that produces a transformation $T = (R,t)$  between the set of correspondence points while minimizing the distance between them. In the transformation matrix $T$, $R$ represents the $3$x$3$ rotation matrix, and $t$ represents the $3$x$1$ translation vector. Finally, the registration transformation is applied to the recorded EM path, registering the two systems in the same coordinate space, $p^m_c, p^m_{em} \in \mathbb{P}^3_{intraop}$, which represents the intraoperative 3D coordinate space.

\section{Evaluation}
\label{sec:Evaluation}

In this study, the introduced DTW method for registration of EM-guided MIEP is evaluated using the mean registration error criterion. The method is compared to path-based ICP registration methods from state-of-the-art. To evaluate the DTW method for registration accuracy, we conducted experiments by recording EM-tracked paths while navigating through each of the six branches of the phantom five times. For each run, the EM catheter-shaped sensor was manually pulled from the main inlet of the phantom towards the outlets of each branch, with an average speed of $1$-$2$cm/s. Subsequently, the recorded EM paths were registered to the phantom's centerline using the introduced DTW method. 

The ground truth registration used in this study is a marker-based method, which is considered a benchmark in the literature. Ten unique easily-identifiable landmarks throughout the phantom are used to register the preoperative model to the EM tracking system for calculating the ground truth. 

\subsection{Mean registration error}
\label{sec:meanregistrationerror}

This criterion is employed to assess the accuracy of the introduced DTW method compared to the ground truth registration. The mean registration error is computed by summing the euclidean distance of each DTW-registered EM path point to its closest point in the ground truth EM-registered path. Equation (\ref{eq:meanerror}) represents the mathematical expression employed for this computation.

\begin{equation}
\label{eq:meanerror}
    \text{Mean error} = \frac{1}{m}\sum_{i=1}^{m}\min_{j}||p^i_{em} - p^j_{gt}||
\end{equation}

The $p^i_{em}$ represents an EM path point transformed by the introduced DTW registration method, $p^j_{gt}$ represents the closest point from ground truth to $p^i_{em}$, $m$ represents the total number of points in both signals, and $||\cdot||$ represents the Euclidean norm. The results of this evaluation criterion are presented in Fig.~\ref{fig:meanerror}.

\section{Results and Discussion}
\label{sec:resultsdicsussion}

\begin{figure*}[htp]
\centering
    \subfloat[\label{fig:dtwresults}]{\includegraphics[width=0.49\textwidth]{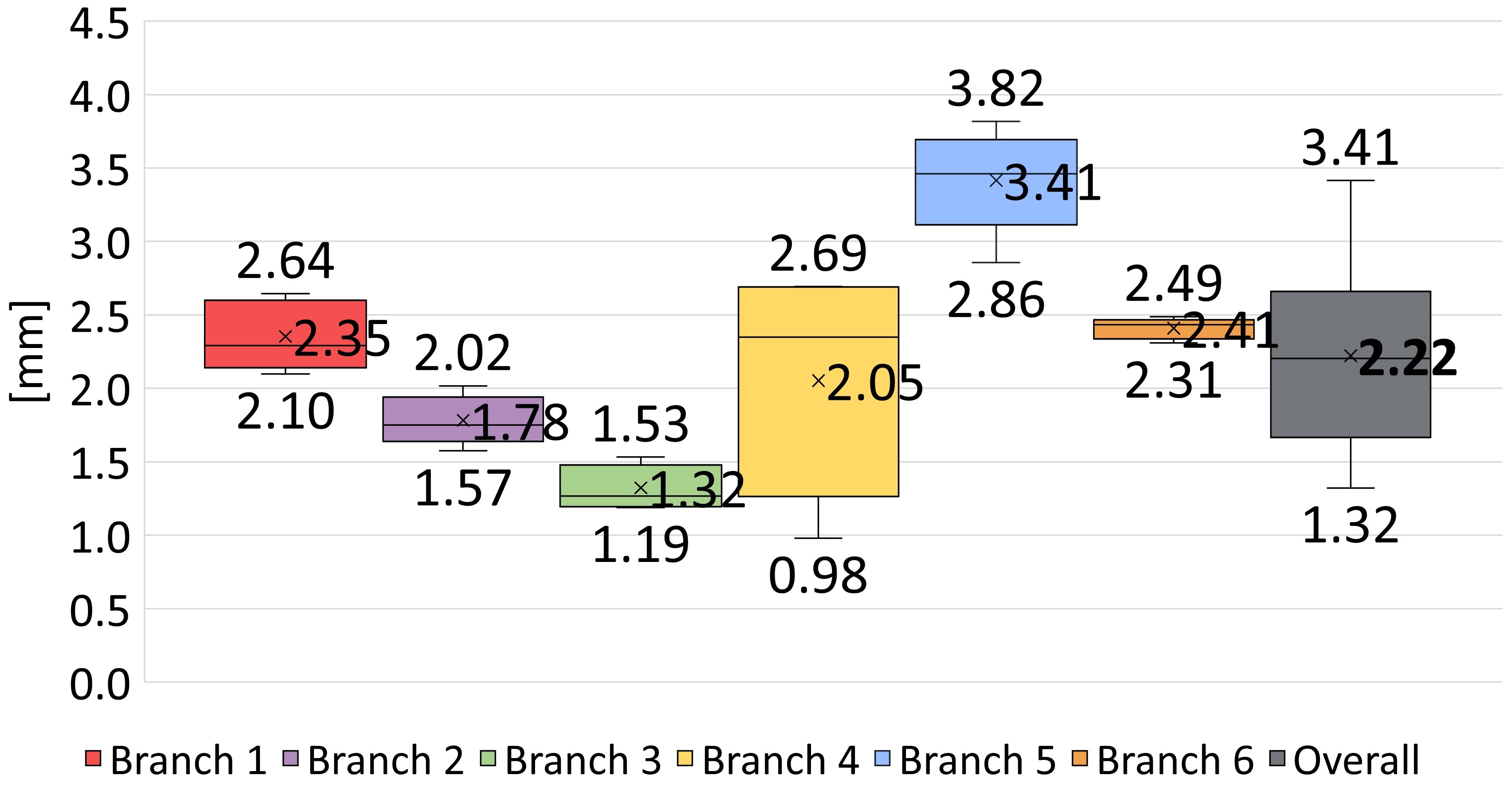}}
    \subfloat[\label{fig:icpresults}]{\includegraphics[width=0.49\textwidth]{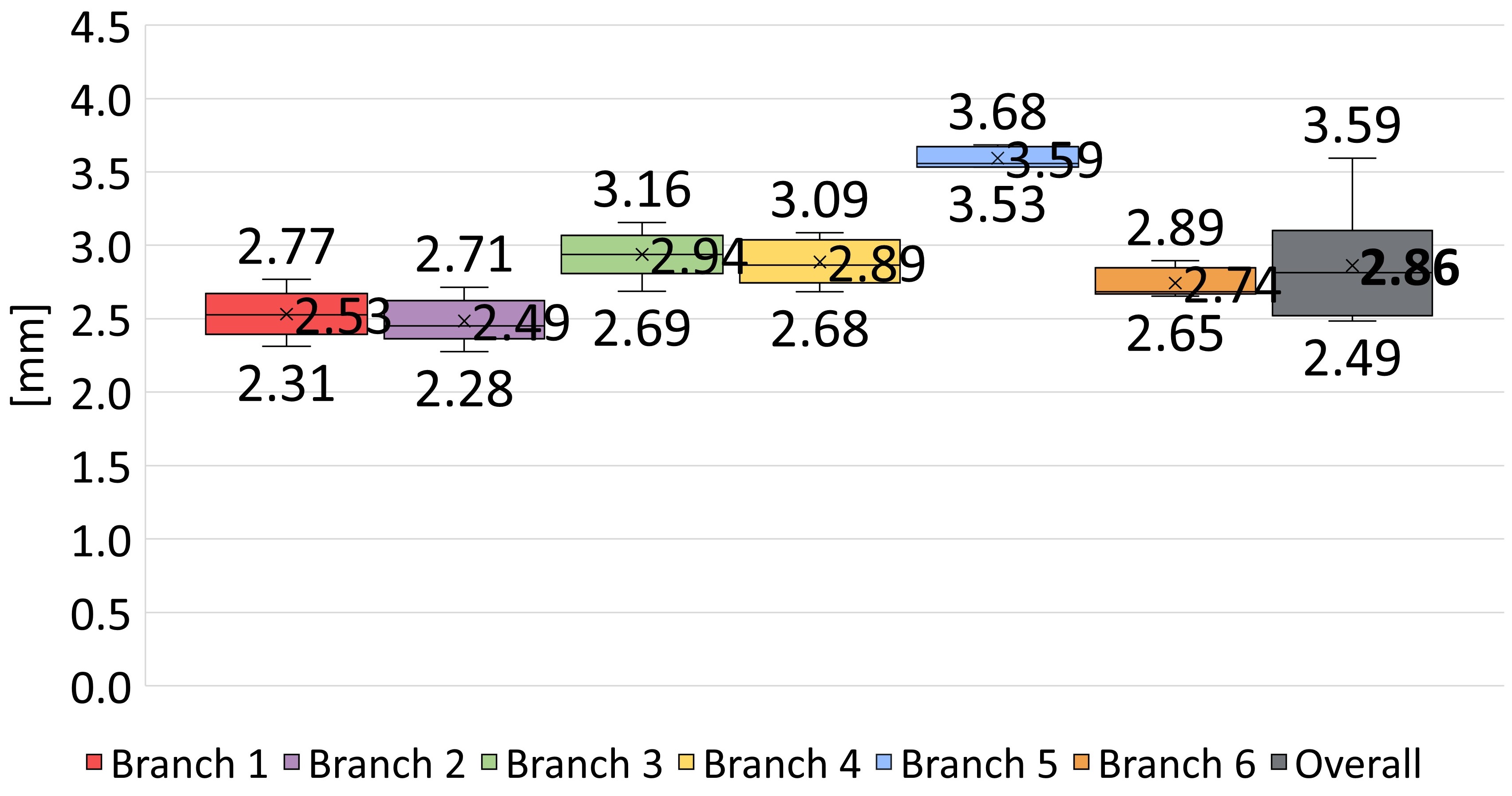}}
    \caption{Mean registration error in millimeters across all branches, and overall accuracy (bars represent the standard deviation, while whiskers indicate the minimum and maximum error). (a) DTW registration of EM paths to centerlines, and (b) ICP registration of EM paths to centerlines.}
\label{fig:meanerror}
\end{figure*}

Based on the experimental setup and evaluation criterion mentioned above, the results of this proof-of-concept study are presented in detail in Fig.~\ref{fig:meanerror}. The mean registration error over all branches and individual runs is $2.22$mm, which falls well within the clinically acceptable range of $<5$mm, as reported in~\cite{nypan_vessel-based_2019}. The proposed DTW registration method performed slightly better than the path-based ICP registration method, with a mean registration error of $2.86$mm.

The results presented in this paper demonstrate that the introduced DTW registration method achieves accurate and reliable registration. The method outperforms the established path-based registration methods that use ICP in all reported branch results. The variability of registration accuracy among different runs is higher in mostly straight vessels, when the automatic selection of registration feature points follows a straight line. This variability is mostly noticeable in results from Branch 4 and 5, where the translation is correctly matched, but the orientation of the signals is not perfectly aligned, and the standard deviation in the results exceed that of ICP.

The introduced methods offers the advantage of automation in the registration process, with no changes in the intraoperative workflow. Unlike marker-based registration methods, which require manual interactions to match point correspondences, the DTW method automatically warps the signals, selects corresponding points, and performs registration. Additionally, the DTW method is not dependent on initialization compared to other path-based registration methods, which may fail to provide a transformation if not correctly initialized. In this paper, the ICP method was consistently initialized from the registered position of the DTW method. This step was necessary as the direct application of ICP registration would not converge to provide a transformation due to the significant distance between the signals.

Furthermore, unlike other path-based registration methods, the DTW registration method does not rely on registering the entire EM path to the centerline. In ICP-based methods, the aim is to minimize the difference between all points to be registered, which means that any deformations in one part of the signal would affect the entire registration. In contrast, the DTW method matches points between signals beforehand and employs algorithms to check the confidence of the matched correspondences, ensuring that the set of points to be registered would result in a reliable registration.

In comparison to prior studies, the proposed method aligns well with other state-of-the-art approaches in the research community. Marker-based techniques outlined in~\cite{manstad-hulaas_three-dimensional_2011} report registration errors of $1.28$mm in phantom studies and $4.18$mm in in-vivo studies. Similarly, two additional studies employ path-based registration methods with ICP and report registration accuracies of $3.75$mm and $4.50$mm in~\cite{nypan_vessel-based_2019} and~\cite{luo_bronchoscopic_2014} respectively. Another ICP registration approach proposed in~\cite{de_lambert_electromagnetic_2012} reports a mean registration accuracy of $1.30$mm. While these comparisons are relative due to differences in tracking data utilized, they demonstrate acceptable registration accuracy, which the introduced DTW method achieves.

Future research directions for the introduced DTW registration include conducting additional evaluations with various phantoms and potentially in-vivo studies, which may provide more evidence of the method's effectiveness in clinical settings. In addition to exploring the impact of catheter movements on DTW matching, another research direction involves improving the approach to registering EM paths to centerlines. In time-dependent series, alternating forward or backward catheter movement can change the signal appearance and result in incorrect matching. To overcome this, one solution is to use the 3D localization and motion-capturing abilities of EM to detect forward and backward movements and then backward warp the signal when the catheter direction changes. Last, advanced algorithms that more accurately depict catheter motion dynamics could be implemented to improve registration accuracy. Current solutions still exhibit significant variability between ground truth catheter movement and the centerline.

\section{Conclusion}
\label{sec:Conclusion}

Accurate catheter tracking is essential in MIEP, and this paper introduces a novel catheter registration method using DTW. As far as the authors know, this study represents the first attempt at using temporal analysis of 3D EM signals for catheter registration. The method is evaluated on a vascular phantom with marker-based registration as the ground truth. The results indicate that the DTW registration method achieves accurate and reliable registration, outperforming path-based ICP registration methods. Furthermore, it provides several advantages compared to existing solutions, including high registration accuracy, registration process automation, preservation of procedural workflow, and elimination of the need for initialization. The method introduced in this paper is a proof-of-concept study, and further experiments by the research community are necessary to establish its applicability and effectiveness in clinical settings. Overall, the introduced DTW registration method has the potential to enhance the accuracy and reliability of catheter tracking in MIEP.

\section*{Acknowledgments}
\label{acknowledgments}
The project was funded by the Bavarian State Ministry of Science and Arts within the framework of the \textit{``Digitaler Herz-OP''} project under grant number 1530/891 02. We also thank ImFusion GmbH and BrainLab AG for their software support and valuable interactions.

%
%
\bibliographystyle{splncs04}
\bibliography{bibliography}

\end{document}